\documentclass{article}
\usepackage{spconf,amsmath,graphicx,multicol}

\def\I{{\bf I}}

\def\w{{\bf w}}

\def\a{{\bf a}}

\def\p{{\bf p}}
\def\r{{\bf r}}
\def\A{{\bf A}}
\def\b{{\bf b}}

\def\x{{\bf x}}
\def\b{{\bf b}}
\def\R{{\bf R}}
\def\sgn{{\mbox {sgn}}}
\def\be{\begin{eqnarray}}
\def\ee{\end{eqnarray}}
\def\Pr{{\rm Pr}}

\def\bmu{\mbox{\boldmath{$\mu$}}}
\def\bdel{\mbox{\boldmath{$\delta$}}}

\def\x{{\mathbf x}}

\title{Why ReLU Units Sometimes Die:  \\
Analysis of Single-Unit Error Backpropagation in Neural Networks}
\name{Scott C. Douglas and Jiutian Yu}
\address{Southern Methodist University  \\
 Department of Electrical and Computer Engineering \\
 Dallas, Texas 75275 USA \\
{\tt sdouglas@smu.edu, jiutiany@smu.edu}}
\begin{document}
%\ninept
%
\maketitle
\begin{abstract}
Recently, neural networks in machine learning use rectified linear units (ReLUs) in early processing layers for better performance. Training these structures sometimes results in ``dying ReLU units'' with near-zero outputs.   We first explore this condition via simulation using the CIFAR-10 dataset and variants of two popular convolutive neural network architectures.  Our explorations show that the output activation probability $\Pr[y>0]$ is generally less than 0.5 at system convergence for layers that do not employ skip connections, and this activation probability tends to decrease as one progresses from input layer to output layer.  Employing a simplified model of a single ReLU unit trained by a variant of error backpropagation, we then perform a statistical convergence analysis to explore the model's evolutionary behavior.  Our analysis describes the potentially-slower convergence speeds of dying ReLU units, and this issue can occur 
 regardless of how the weights are initialized. 
\end{abstract}
%\vspace{-.07in}

\vspace{-.07in}
\section{Introduction}
\vspace{-.05in}

Machine learning has seen a recent explosion in both research and development due to the availability of large data sets, useful adaptive algorithms and architectures, and fast computational processing for training.  One popular class of architectures are multilayer neural networks that uses rectified linear units, or ReLUs, in the early stages of processing, in which a single-unit input-output relation within the network is 
\be
y & = & \w^T \x  + b \\
f(y) & = & \max(0, y),
\ee
where $\w = [w_1 \; \cdots \; w_N]^T$ is the weight vector for the particular unit, 
$\x = [x_1 \; \cdots \; x_N]^T$ is the unit's input vector, $b$ is the bias weight, and $f(y)$ is the ReLU activation function
\cite{Rumelhart1986a,Hinton2010,Bengio2012,Zeiler2013}.  The ReLU activation function is simple to compute.  It also avoids difficulties with regard to output scaling. 

Systems employing these units sometime exhibit an issue during training known as ``dying ReLU units,''  in which the condition $y < 0$  is highly likely for almost all training patterns for many units within the network. 
Little is understood about this condition other than the simple observation that the error backpropagation algorithm employs a function $f'(y)$ that for the ReLU nonlinearity is zero for $y<0$, such that patterns that cause $y<0$ do not change the unit's weights.  It is suggested in \cite{Ng2013} that this condition could be problematic in practice, because units that are not active will likely not be trained, and thus ``leaky ReLUs,'' that make $f'(y)>0$ for all $y$ are introduced.     However, the results in \cite{Ng2013} using such modified activation functions are not substantially-different from those using ReLUs.   To our knowledge, the issue of ``dying ReLUs'' has not been extensively explored in the machine learning literature.  

This paper explores the issue of ``dying ReLUs'' in two ways.  First, we perform extensive simulations of multiple variants of the VGG and ResNet architectures on the CIFAR-10 image classification dataset to numerically evaluate the output activation probability $\Pr[y>0]$ across each layer of each structure for both training and test data.  Our results produce the following observations:   
\begin{enumerate}
\item  The average value of $\Pr[y>0]$ generally decreases as one progresses from input layer to output layer for those layers that do not have direct skip connections. 
\item  The convergence speed of the value of $\Pr[y>0]$ is generally slower for layers that exhibit small values of $\Pr[y>0]$ near convergence.    Thus, units that are ``dying'' appear to be slower to converge.
\end{enumerate}
Second, to understand the convergence speed issue associated with varying $\Pr[y>0]$ unit values, we analytically explore the convergence behavior of a simple single-unit $(L+1)$-parameter (signal plus bias weights)  error backpropagation model given by the update relations
\be
\w_{new} & = & \w + \eta  (d - f(y)) f'(y)  \x
				\label{eq:ReLU1}
 \\
b_{new} & = & b + \eta (d- f(y)) f'(y) 
				\label{eq:ReLU2}
\ee
where $\w_{new}$ and $b_{new}$ are the updated weights,  $d$ is the target or training signal for the unit, and $u(y) = f'(y)$ is the unit step function, under a simple statistical model for $d$ and $\x$.  The approach taken follows that employed for understanding linear adaptive filters with non-quadratic error costs as well as nonlinear adaptive systems such as the single-layer error backpropagation algorithm with saturating nonlinearities
\cite{Bershad1993a,Bershad1993b,Douglas1994}.  Our statistical analysis shows that the combined mean weight vector $[ E\{ \w^T \}  \; E\{b\}]^T$
converges approximately exponentially near its optimal solution with all but three of its modes of the form $(1 - \eta \Pr[y>0])^k$.  
Moreover, these converging modes are likely to dominate the overall convergence rate of the adaptive parameters due to both their multiplicity and their slow rate when $\Pr[y>0]$ is small.  
Thus, the issue of ``dying ReLUs'' is one of a potentially-slow convergence rate due to their sparse output activations, as opposed to the absolute sparsity of the output activations themselves.   Moreover, unlike the concerns raised in 
\cite{Ng2013}, this issue appears not to be tied to how the units are initialized.  
\vspace{-.07in}
\section{Numerical Evaluations}
\vspace{-.05in}

In order to better understand the underlying convergence issues, we first experiment using state-of-the-art architectures from the very deep convolutional networks (VGG)
\cite{Simonyan2015} and residual networks (ResNet) 
\cite{He2016} families; specifically, VGG-9, VGG-16, ResNet-20, and ResNet-32.  VGG-9 is a cropped version of VGG-16 containing 7 convolutional layers followed by two fully-connected layers.  A batch normalization layer is added after each convolutional layer and the first fully-connected layer.  We apply these systems to a popular image classification benchmark:  CIFAR-10, which consists of $(32 \times 32)$-pixel color images across ten classes, with 50000 images for training and 10000 images for testing
\cite{Alex2009}.
For each architecture, we calculate the probability of non-zero outputs of the ReLU units at each layer at each epoch during training; similar values were also observed on the testing data.  To improve classification performance, flipping and translation data augmentation has been used.    We employ standard gradient descent for each model with the cross-entropy loss, with the following parameters:  a mini-batch size of 128, weight decay of 0.0001, a momentum value of 0.9, and He normal weight initialization.  As for learning rates, values were chosen to be initially large for the VGG and ResNet architectures and then reduced by a factor of 1/10 at epochs 150 and 100, respectively, and again at epochs 225 and 200, respectively, except for a short 10-epoch initial training period for ResNet whereby a small step size was used.  
Averages across ten different training runs have been computed in each case.  
\begin{figure}[t]
\vspace{-1.3in}
\hspace{-.4in}
\includegraphics[width=10.5cm]{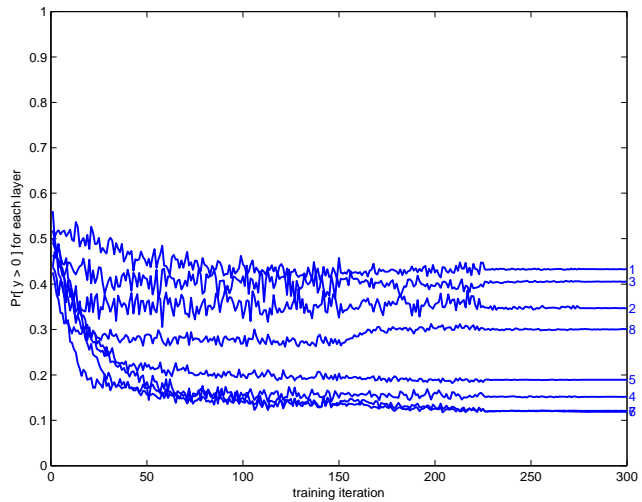}

\vspace{-1.5in}
%\vspace{2.5in}
\caption{Evolution of $\Pr[y>0]$ values for the VGG-9 architecture on CIFAR-10.}
\vspace{-0.2in}
\end{figure}

\begin{figure}[t]
\vspace{-1.3in}
\hspace{-.4in}
\includegraphics[width=10.5cm]{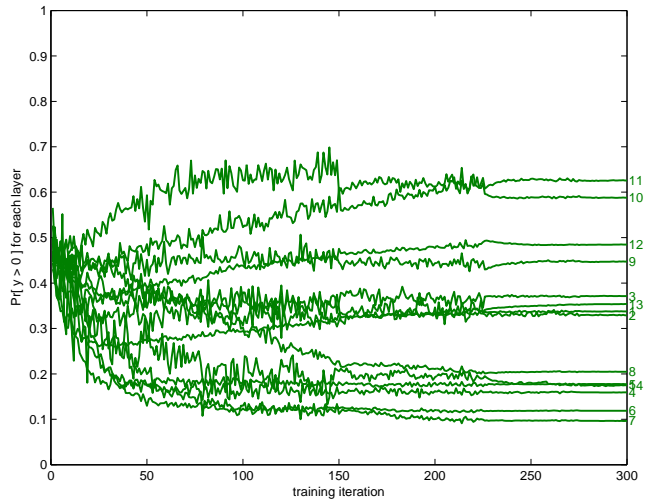}

\vspace{-1.5in}
%\vspace{2.5in}
\caption{Evolution of $\Pr[y>0]$ values for the VGG-16 architecture on CIFAR-10.}
\vspace{-0.2in}
\end{figure}

Figure~1 shows the evolutions of $\Pr[y>0]$ for the various layers of VGG-9, where it is seen that the output activation probabilities tends to decrease during training for all of the layers.  There is a tendency for layers that have lower $\Pr[y>0]$ to have a somewhat slower convergence of the observed $\Pr[y>0]$ values near convergence, which upon further inspection is particularly evident for Layers 5, 6, and 7 prior to the first step size change at epoch 150.  All of the layers have a reduced (less than 0.5) output activation probability at convergence, indicating that output sparsity is a desirable trait from the standpoint of classification performance with this architecture.  

Figure~2 shows the evolutions of the output activation probabilities for VGG-16, for which the trends are not as clear.   Considering only the first training period up to epoch 150, the slowest-converging layer appears to be Layer 8 in the middle of the architecture, which also has a low $\Pr[y>0] \approx 0.204$ value at convergence.    Layers after this layer, however, also exhibit somewhat slower convergence to higher final $\Pr[y>0]$ values after having smaller values at intermediate points in training.  

Figures~3 and~4 show the evolutions of $\Pr[y>0]$ values for the two ResNet architectures, respectively, where it is seen that layers with skip connections, shown in red, generally have higher $\Pr[y>0]$ values throughout training.  This is to be expected given the nature of skip connections, which are designed to increase the possibility of activation where they are used.  The layers that have lower steady-state 
$\Pr[y>0]$ values are those without skip connections, shown in blue and green, respectively, which also tend to converge more slowly over the fastest training period from epoch 10 to epoch 100.  In this case, Layer 15 in the ResNet-20 architecture also exhibits slow convergence after experiencing a significant change to a low $\Pr[y>0]$ value in the initial training period.  

Figure~5 shows the average values of $\Pr[y>0]$ obtained for each layer at convergence for the VGG (top) and ResNet (bottom) architectures in our evaluations.  An interesting behavior is noted:  The output activation probabilities decrease from input to output for VGG-9 and both ResNet architectures when considering the presence or absence of skip connections.   This characteristic was not exhibited by VGG-16, which had its lowest $\Pr[y>0]$ values near the middle of the architecture.  
\vspace{-.11in}

\begin{figure}[t]
\vspace{-1.3in}
\hspace{-.4in}
\includegraphics[width=10.5cm]{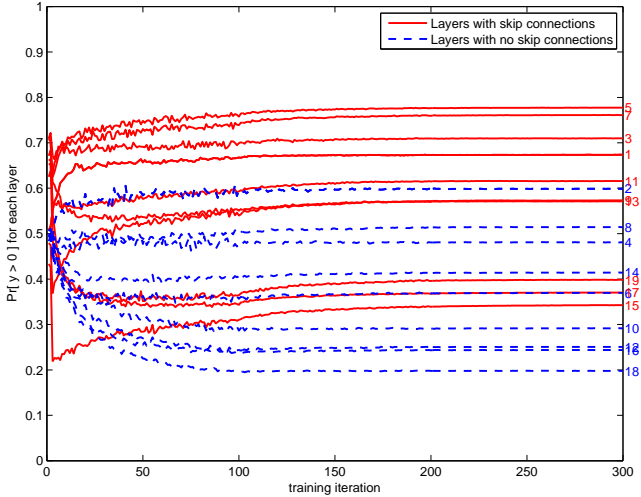}

\vspace{-1.5in}
%\vspace{2.5in}
\caption{Evolution of $\Pr[y>0]$ values for the ResNet-20 architecture on CIFAR-10.}
\vspace{-0.2in}
\end{figure}

\begin{figure}[t]
\vspace{-1.3in}
\hspace{-.4in}
\includegraphics[width=10.5cm]{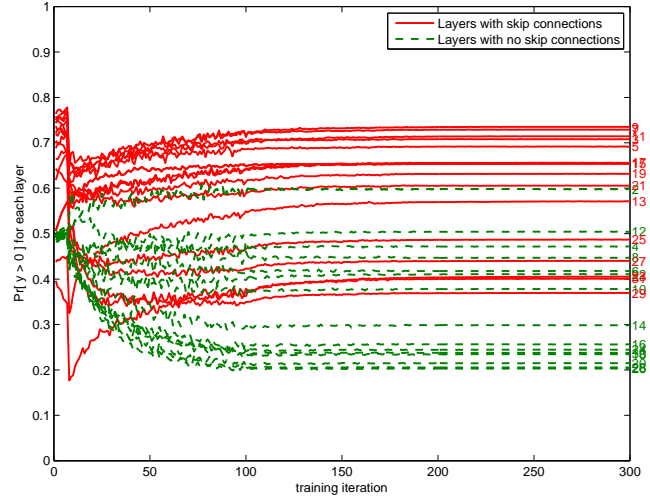}

\vspace{-1.5in}
%\vspace{2.5in}
\caption{Evolution of $\Pr[y>0]$ values for the ResNet-32 architecture on CIFAR-10.}
\vspace{-0.2in}
\end{figure}

\vspace{-.07in}
\section{Statistical Analysis}
\vspace{-.05in}

In this section, we develop an analytical model to explore features of the convergence behavior observed in the previous section.  The model chosen is quite simple and assumes that any particular unit within the network evolves according to a single-unit model of the form in (\ref{eq:ReLU1})--(\ref{eq:ReLU2}), where the input signal vector follows a Gaussian distribution 
with mean vector $\bmu$ and an identity covariance matrix $\I$, or 
\be
\x & \sim & {\cal N}( \bmu, \I ).  
\ee
Furthermore, we assume that the desired response follows a linear model of the form
\be
d & = & \a^T \x + \overline{c}, 
\ee
where the vector $\a$ and $\overline{c}$ contain unknown parameters.  

Our analysis employs a locally-convergent variant of the update in (\ref{eq:ReLU1})--(\ref{eq:ReLU2}), given by
\be
\w_{new} & = & \w + \eta  (d - y) u(d)  \x   \label{eq:ReLUmodel1} \\
b_{new} & = & b + \eta (d - y) u(d),  \label{eq:ReLUmodel2}
\ee
where  $u(y)$ is the unit step function, which can be shown to be equivalent to (\ref{eq:ReLU1})--(\ref{eq:ReLU2}) when $y$ is in the close vicinity of $d$.    Eqns. (\ref{eq:ReLUmodel1})--(\ref{eq:ReLUmodel2}) are identical to the well-known least-mean-square (LMS) adaptive filtering algorithm with data-dependent step size $\eta(d) = \eta u(d)$, admitting a straightforward dynamic analysis.  
%It is straightforward to show for the ReLU nonlinearity that the above update is equivalent to 
%\be
%\w_{new} & = & \w + \eta  (d - y) u(y)  \x \\
%b_{new} & = & b + \eta (d- y) u(y), 
%\ee
%where $u(y)$ is the unit step function.  Our analysis employs the variant of the above update given by
%\be
%\w_{new} & = & \w + \eta  (d - y) u(d)  \x \\
%b_{new} & = & b + \eta (d - y) u(d),
%\ee
%which, for $d$ close to $y$ is 
%\be
%\w_{new} & = & \w + \eta ( d - y ) f'(y)  \x, 
%\ee
%where
%\be
%y & = & \w^T \x  + b \\
%f(y) & = &\mbox{max}[y,0] \\
%& = &  \left\{ \begin{array}{cc}
%y & y > 0 \\
%0 & y \leq 0 
%\end{array} \right. 
%\ee
%The derivative of the ReLU nonlinearity is
%\be
%f'(y) & = & u(y) \\
%& = &  \left\{ \begin{array}{cc}
%1 & y > 0 \\
%0 & y \leq 0 
%\end{array} \right. 
%\ee
%Therefore,
%\be
%\w_{new} & = & \w + \eta(y) (d - y) \x
%\ee
%where
%\be
%\eta(y) & = & \eta u(y)
%\ee
%
%
%We approximate the update using the following:  
%\be
%\w_{new} & = & \w + \eta  (d - y) u(d)  \x \\
%b_{new} & = & b + \eta (d - y) u(d) \\
%y & = & \w^T \x + b
%\ee
%This relation should closely follow the original one when $d \approx y$, which will be the case near convergence.

Without loss of generality, let $\a$ take the simplified form
\be
\a & = &  \left\{ \begin{array}{cc}
a & i=L \\
0 & 1 \leq i \leq L-1. 
\end{array} \right.
\ee
This form is valid because of the covariance structure of $\x$, which is rotation-invariant; hence, we can rotate the axes of the analysis to allow the above form.
Then,
\be
d & = & a x_L + c \\
x_L & = & \frac{d - c}{a}. 
\ee
%where 
%\be
%E\{ d \} %& = & \a^T \bmu_{orig} + \overline{c} \\
%& = & a \mu_L + c
%\ee
Thus, the desired response signal is also Gaussian with distribution
\be
d & \sim & {\cal N}(\mu_d, a^2), \;\;\;\; 
\mu_d =  a \mu_L + c. %\\
%d & \sim & {\cal N}(a \mu_L +c, a^2) 
\ee
\begin{figure}[t]
\vspace{-1.3in}
\hspace{-.4in}
\includegraphics[width=10.5cm]{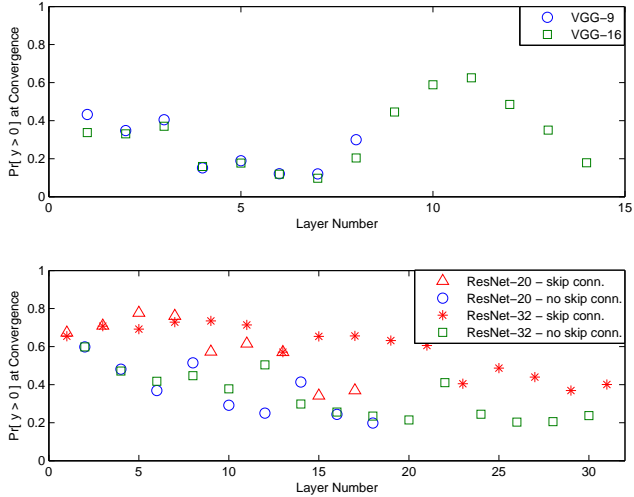}

\vspace{-1.5in}
%\vspace{2.5in}
\caption{Final $\Pr[y>0]$ values for the VGG (top) and ResNet (bottom) architectures on CIFAR-10.}
\vspace{-0.2in}
\end{figure}
This signal model allows us to write the coefficient updates of the algorithm analysis model as
\be
\!\!\!\! \w_{new} & \!\!\! = \!\!\! & \w + \eta u( d)\!\! \left(\!\!  d - w_L \frac{d - c}{a} - b -  \sum_{j=1}^{L-1} w_j x_j\!\!   \right) \!\!  \x \\
\!\!\!\!  b_{new} & \!\!\!  = \!\!\! & b + \eta u(d) \!\! \left(\!\!    d - w_L \frac{d - c}{a}   - b -  \sum_{j=1}^{L-1} w_j x_j \!\!  \right). 
\ee

At this point, we are able to determine the evolutionary equations for the mean coefficient vector $E\{ \w \}$ and bias $E\{ b \}$ under the independence assumptions \cite{Mazo1972} as is done in linear adaptive systems.  For this analysis, we require the quantities
\be
E\{ d^n u(d) \} & \!\!\! = \!\!\!  & \int_0^\infty \!\!\! \frac{x^n}{\sqrt{2 \pi a^2}} \exp \left( - \frac{(x - \mu_d)^2}{2 a^2} \right) dx. 
\ee
It can be shown that
\be
E\{ u(d) \} & = & \Phi \left( \frac{\mu_d}{|a|} \right) \\
& = &  \frac{1}{2} \left( \mbox{erf} \left( \frac{\mu_{d}}{\sqrt{2} |a|} \right) + 1 \right) \\
E\{ d u(d) \} & = & \frac{|a|}{\sqrt{2 \pi}} \exp \!\!\left( \!\!  - \frac{\mu_d^2}{2 a^2}\!\!  \right) \! + \mu_d E\{ u(d) \}
\ee
\be
\!\!\!\!\!\! E\{ d^2 u(d) \} & \!\!\! = \!\!\! & \frac{|a| \mu_d }{\sqrt{2 \pi}} \exp \!\!\left( \!\! - \frac{\mu_d^2}{2 a^2} \!\!  \right) \! + (\mu_d^2 + a^2)  E\{ u(d) \}
\ee

With the above results, we have
\be
E\{ \w_{new} \} & = & \left( \I - \eta \R \right) E\{ \w \}  - \eta \r E\{b\} + \eta \p  \\
E\{ b_{new} \} & = & \left( 1 - \eta E\{ u(d) \} \right) E\{ b\} -  \eta \r^T E\{ \w \}  \nonumber \\
& & + \eta E\{ d u(d) \}, 
\ee
where we have defined 
\be
\R & = & E\{ u(d)  \x \x^T \}  \\
\r & = & E\{ u(d) \x \} \\
\p & = & E\{ d u(d) \x \}. 
\ee
The entries for $\R$, $\r$, and $\p$ can be evaluated using the formulas for $E\{ d^n u(d)\}$ shown previously, the details of which are omitted for brevity.  The expressions obtained are shown at the top of the next page.  
\begin{table*}[th]
\be
[\R]_{ij} & \!\!\! = \!\!\!& \left\{ \begin{array}{rc}
( \mu_i \mu_j  + \delta_{ij} ) E\{ u(d) \} & 1 \leq i \leq L-1 \mbox{  and  } 1 \leq j \leq L-1 \\
\frac{1}{\sqrt{2 \pi}} 
\mu_i \sgn(a) \exp \left( - \frac{\mu_d^2}{2 a^2} \right) + \mu_i \mu_L E\{ u(d) \} 
& 1 \leq i \leq L-1 \mbox{  and  } j = L  \\
\frac{1}{\sqrt{2 \pi}} 
\mu_j \sgn(a) \exp \left( - \frac{\mu_d^2}{2 a^2} \right) + \mu_L \mu_j E\{ u(d) \}
& i = L \mbox{  and  } 1 \leq j \leq L-1  \\
\frac{1}{\sqrt{2 \pi}} \left( \mu_L \sgn(a) - \frac{c}{|a|} \right) 
\exp \left( - \frac{\mu_d^2}{2 a^2} \right) + ( \mu_L^2 + 1) E\{ u(d) \}
& i = j =  L
\end{array} \right.
\ee
\vspace{-.1in}
%\setcounter{equation}{15}
%
%\be
%[\p]_i & = & \left\{ \begin{array}{rc}
%a \mu_i \left(  \frac{1}{\sqrt{2 \pi}} \sgn(a)  \exp \left( - \frac{\mu_d^2}{2 a^2} \right) +  (\mu_L + \frac{c}{a})  E\{ u(d) \} \right)
%& 1 \leq i \leq L-1 \\
%a \left(\frac{1}{\sqrt{2 \pi}} \mu_L \sgn(a)  \exp \left( - \frac{\mu_d^2}{2 a^2} \right) + (\mu_L^2 + 1 + \frac{c}{a} \mu_L)  E\{ u(d) \} \right)
%& i = L
%\end{array} \right.
%\ee
\be
[\p]_i & = & \left\{ \begin{array}{rc} a [ \R]_{iL} + \mu_i c  E\{ u(d) \},  &   1 \leq i \leq L-1 \\
a \left(\frac{1}{\sqrt{2 \pi}} \mu_L \sgn(a)  \exp \left( - \frac{\mu_d^2}{2 a^2} \right) + (\mu_L^2 + 1 + \frac{c}{a} \mu_L)  E\{ u(d) \} \right)
& i = L
\end{array} \right.
\ee
\vspace{-.1in}
\be
[\r]_i & = & \left\{ \begin{array}{rc}
\mu_i E\{ u(d) \} 
& 1 \leq i \leq L-1 \\
\frac{1}{\sqrt{2 \pi}} \sgn (a)   \exp \left( - \frac{\mu_d^2}{2 a^2} \right) + \mu_L E\{ u(d) \}  
& i = L
\end{array} \right.
\ee
\vspace{-.4in}
\end{table*}

We can combine the updates for both $\w$ and $b$ in the $(L+1)$-element vector
\be
\overline{\w} & = & \left[ \begin{array}{c}
b \\
\w
\end{array} \right]. 
\ee
Let us define 
\be
\overline{\bmu} & = & \left[ % \begin{array}{c}
1 \;  \;  %\\
\mu_1  \; \; %\\
\cdots \;  \; % \vdots \\
\mu_{L-1} \;  \; %\\
h
%\end{array}
 \right]^T
\ee
where 
\be
h & = & \mu_L +  \frac{\sgn (a)}{\sqrt{2 \pi} E\{ u(d) \}}    \exp \left( - \frac{\mu_d^2}{2 a^2} \right). 
\ee
Then, the evolutionary equation for the means of the elements of $\overline{\w}$ is 
\be
E\{ \overline{\w}_{new} \} & = & ( \overline{\I} - \eta  \A) E\{ \overline{\w}  \}+ \eta   \b, 
\ee
where
\be
\A & \!\!\! = \!\!\! & E\{ u(d) \}   ( \overline{\I} + \overline{\bmu} \overline{\bmu}^T  - \overline{\bdel}_1 \overline{\bdel}_1^T)  - K\overline{\bdel}_{L+1} 
\overline{\bdel}_{L+1}^T   \\
\b & \!\!\! = \!\!\! &  \overline{\bmu} E\{ u(d) \}   (c + a h)   +  \overline{\bdel}_{L+1}  a ( E\{ u(d) \} - K) \\
%K & \!\!\! = \!\!\! &  E\{ u(d) \}(1 +   h^2) - [ \R]_{LL}  \\
%K & \!\!\! = \!\!\! &  - \frac{1}{\sqrt{2 \pi}} \left( \mu_L \sgn(a) - \frac{c}{|a|} \right) 
%\exp \left( - \frac{\mu_d^2}{2 a^2} \right) - ( \mu_L^2 + 1) E\{ u(d) \}
%+ E\{ u(d) \} +  E\{ u(d) \}  \left( \mu_L +  \frac{\sgn (a)}{\sqrt{2 \pi} E\{ u(d) \}}    \exp \left( - \frac{\mu_d^2}{2 a^2} \right) \right)^2  \\
K & \!\!\! = \!\!\! &  \frac{1}{\sqrt{2 \pi}} \left( \mu_L \sgn(a) + \frac{c}{|a|} \right) 
\exp \left( - \frac{\mu_d^2}{2 a^2} \right)  \nonumber \\
& & 
+  \frac{1}{2 \pi E\{ u(d) \}} \exp \left( - \frac{\mu_d^2}{a^2} \right). 
\ee 

Examining the form of $\A$ above, we see that it has $(L-2)$ eigenvalues equal to $E\{ u(d) \}  = \Pr[d > 0] \approx \Pr[y>0]$ for our assumed analytical model.  Thus, convergence in almost all signal dimensions will be exponential with the factor 
$(1 - \eta \Pr[y>0])$ near convergence.  It is generally not possible to choose a large step size to obtain faster convergence, as other signal dimensions have larger eigenvalues which limit the maximum step size allowed.   Thus, if $\Pr[y>0]$ is small, the unit will have slow convergence in these dimensions.

\begin{figure}[t]
\vspace{-2in}
\hspace{-0.9in}
\includegraphics[width=13cm]{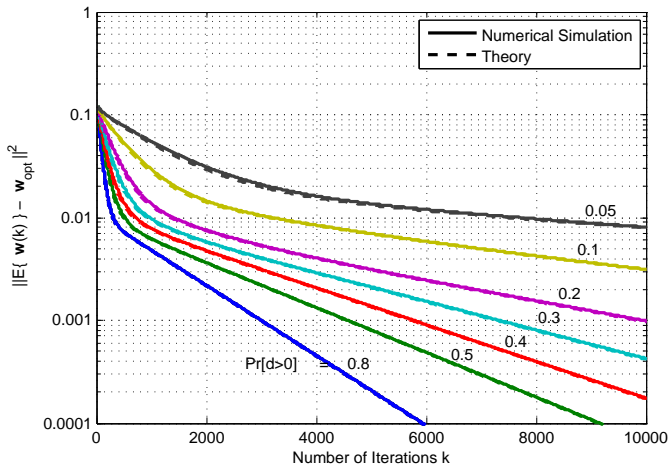}

\vspace{-2.2in}
%\vspace{2.5in}
\caption{Analysis model performance verification; see text for explanation.}
\vspace{-0.2in}
\end{figure}

We have performed Monte Carlo simulations to verify the equations derived.  In these simulations, we have implemented an $L=11$-parameter single ReLU adaptive unit updated using the analysis model  in (\ref{eq:ReLUmodel1})--(\ref{eq:ReLUmodel2}) with $\eta u(d)$ step size, Gaussian input and desired response signals with $\bmu  = [2 \; 1.6 \; 1.2 \; \cdots \; -1.6 \; -2.0]^T$, $a=0.5$, and $c$ chosen to obtain $\Pr[d>0]$ values given by one of  $\{ 0.8, 0.5, 0.4, 0.3, 0.2, 0.1, 0.05\}$.  The initial weight errors are distributed at ${\cal N}(0,0.01)$, and the average squared weight error norm is computed across 10000 algorithm iterations and 1000 simulation runs to compare with the theory.  Figure~6 shows the convergence of the algorithm predicted by the theoretical relations and from the numerical simulations, where close agreement is seen.  In addition, it is observed that the convergence speed is slowed as $\Pr[d>0]$ is reduced, which verifies the analytical prediction.  

To verify that the analytical model accurately depicts the behavior of the original single-unit ReLU update in (\ref{eq:ReLU1})--(\ref{eq:ReLU2}) with $\eta u(y)$ step size, we repeat the above simulations for the original algorithm.  Figure~7 shows the results, in which the observed behavior of the original algorithm is quite similar to that predicted by the analysis model near convergence, except that the original algorithm is initially slower for small $\Pr[d>0]$ values. 
\vspace{-.1in}

\begin{figure}[t]
\vspace{-2in}
\hspace{-0.9in}
\includegraphics[width=13cm]{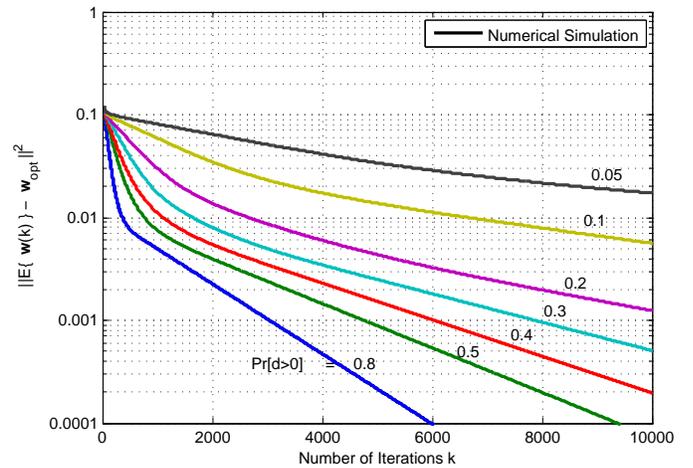}

\vspace{-2.2in}
%\vspace{2.5in}
\caption{Original model performance verification; see text for explanation.}
\vspace{-0.2in}
\end{figure}

\vspace{-.07in}
\section{Conclusions}
\vspace{-.05in}

In this paper, we study issues surrounding ``dying ReLUs'' in deep neural networks.  
Simulation studies applying well-known architectures to the CIFAR-10 benchmark show that output activations within the networks are sparse, particularly for layers with no skip connections.  We also analyze a simple single-unit model for understanding the convergence behaviors of dying ReLU units.  Additional work is ongoing.
\vspace{-.1in}

\begin{center}

\end{center}

\end{document}